\documentclass[lettersize,journal]{IEEEtran}
\usepackage{amsmath,amsfonts}
\usepackage{algorithmic}
\usepackage{algorithm}
\usepackage{array}
\usepackage[caption=false,font=normalsize,labelfont=sf,textfont=sf]{subfig}
\usepackage{textcomp}
\usepackage{stfloats}
\usepackage{url}
\usepackage{verbatim}
\usepackage{graphicx}
\def\BibTeX{{\rm B\kern-.05em{\sc i\kern-.025em b}\kern-.08em
    T\kern-.1667em\lower.7ex\hbox{E}\kern-.125emX}}
\usepackage{balance}

\begin{document}
\title{Real-Time Model-Based Quantitative Ultrasound and Radar}
  
\author{Tom Sharon, Yonina C. Eldar \IEEEmembership{Fellow, IEEE}
\thanks{This project has received funding from the European Research Council (ERC) under the European Union’s Horizon 2020 research and innovation programme (grant agreement No. 101000967), from the Manya Igel Centre for Biomedical Engineering and Signal Processing, and from an amazon fellowship.}
\thanks{T. Sharon and Y. C. Eldar are with the Faculty of Math and Computer Science, Weizmann Institute of Science, Rehovot, Israel. (e-mail:tom.sharon@weizmann.ac.il, yonina.eldar@weizmann.ac.il). }}

\maketitle

\begin{abstract}
Ultrasound and radar signals are highly beneficial for medical imaging as they are non-invasive and non-ionizing. Traditional imaging techniques have limitations in terms of contrast and physical interpretation. Quantitative medical imaging can display various physical properties such as speed of sound, density, conductivity, and relative permittivity. This makes it useful for a wider range of applications, including improving cancer detection, diagnosing fatty liver, and fast stroke imaging. However, current quantitative imaging techniques that estimate physical properties from received signals, such as Full Waveform Inversion, are time-consuming and tend to converge to local minima, making them unsuitable for medical imaging. To address these challenges, we propose a neural network based on the physical model of wave propagation, which defines the relationship between the received signals and physical properties. Our network can reconstruct multiple physical properties in less than one second for complex and realistic scenarios, using data from only eight elements. We demonstrate the effectiveness of our approach for both radar and ultrasound signals.
\end{abstract}


\section{Introduction}
\label{sec:introduction}
\IEEEPARstart{M}{deical}imaging constitutes a non-invasive method to see inside the human body and improve diagnoses, treatment and monitoring of diverse medical conditions. Ultrasound (US) and radar are two primary signals for this purpose allowing non-ionizing, non-invasive, and accessible medical imaging. The image is created based on the received signals, referred to as Channel Data (CD), created from reflections from the medium by the transmitted US or radar signals. Standard imaging is based on beamforming algorithms such as Delay-And-Sum (DAS) beamforming that applies a weighted sum over the receiving signals, after an appropriate delay based on the receiving array geometry \cite{van1988beamforming, lim2008confocal}. These methods often have limited resolution and contrast, and lack physical interpretation. 

Quantitative imaging displays different physical properties of each pixel of the scanned medium. For instance, since malignant tumor cells have higher Speed of Sound (SoS) values than benign cells \cite{ruby2019breast}, a quantitative image of the SoS values of the scanned medium may allow better cancer detection. Additionally, density quantitative imaging is beneficial for fatty liver disease treatment as it allows better quantification of the fat percentage of the liver \cite{lee2017imaging}. Previous studies \cite{hopfer2017electromagnetic,ireland2011microwave, wei2018deep} also highlighted the importance of quantitative imaging of the brain, specifically for cost-effective and fast stroke imaging and classification.

In order to achieve quantitative imaging, a non-linear Inverse Scattering Problem (ISP) needs to be solved which is the problem of determining the characteristics of an object, based on data that is scattered from it. There are known iterative optimization methods, based on gradient descent, such as the Full Waveform Inversion (FWI) and the Nonlinear Waveform Inversion (NWI) for solving the ISP \cite{shultzman2022nonlinear, guasch2020full}. However, these techniques are time-consuming and tend to converge to local minima. In addition, these algorithms often diverge for nonhomogeneous background of the scanned medium (large disparity in the physical values), making them unsuitable for medical imaging \cite{virieux2009overview}.

Recently deep learning methods were suggested for solving the ISP to achieve real-time results and to avoid converging to local minima. To succeed in learning the complex relation between the CD and the physical properties mappings, it was suggested to use model based deep learning approaches which combine a known model in the training or network design \cite{guo2023physics, chen2020review}. Model-based approaches are known to lead to more accurate networks while requiring fewer learned parameters \cite{shlezinger2022model, shlezinger2023model}. To address the ISP, we can incorporate the known model of the wave propagation equation, which calculates the CD given the physical properties of the medium. However, previous works have several limitations which have precluded their adoption in real-time systems, as we explain in the next subsection. Here, we present a neural network based on the physical model of wave propagation that reconstructs in real-time physical properties mappings from either radar or US signals, using only eight elements and for diverse transmission setups.

\subsection{Literature Review}
\noindent Various model-based deep-learning approaches have been used to solve the ISP and can be classified into three categories based on their design \cite{guo2023physics}. The first category consists of networks that enhance the reconstructed images produced by conventional physical methods \cite{li2018deepnis}. The second involves networks that impose physics constraints, such as in the training loss \cite{jin2020physics, liu2021ultrasound}. The third category comprises networks that are designed based on the physics model itself. This type of network benefits from the known physics model, as the network only needs to learn the unknown relationships between physical properties and the CD, rather than learning the physics model from scratch. Examples of these networks include the method proposed in \cite{zhang2020adjoint} which uses a U-Net to reconstruct SoS maps for seismic data. The network's input in each layer includes the SoS estimation map and a set of gradient calculations for each source. However, this approach tends to lead to errors due to implementation approximations of the multiple gradient calculations and incurs long computational time. Furthermore, the network needs 200 elements for the received CD. In \cite{fan2022model} the authors use a CNN to learn the proximal operator in the Primal-Dual Hybrid Gradient (PFHG) method, and a U-Net based frequency-to-image domain network to replace the adjoint operator calculations. However, the network is based on the paraxial approximation which assumes that the elements are around the object in a circle, and therefore are not suitable for different transmission setups such as a linear probe. Transmission setups of elements in different sides or surrounding the object are assumed in most of the works for radar or US signals, which created symmetries in the received signals and made the reconstruction process easier \cite{prasad2022deepuct, rao2020multi, liu2021ultrasound, suo2023data, jin2021deep, wei2018deep, liu2021physical, guo2021physics}. 

Other techniques for solving the ISP involve defining the reconstructed physical property as a learned property in a wave propagation model, such as SWINet for the reconstruction of SoS for seismic data \cite{ren2020physics} or in \cite{hu2021theory} for relative permittivity. These methods may be less stable due to the dependency on the Partial Differential Equation (PDE) in the backpropagation calculations and require a significant number of elements (for instance, 384 receivers' data for SWINet). Moreover, these methods need to be trained for each example, making them unsuitable for real-time imaging.

For radar signals, there are works that used the Supervised Descent Method (SDM) to learn a set of descent directions instead of computing the Fréchet derivative and gradients for each iteration \cite{guo2021physics}. However, these techniques suffer from relatively slow prediction due to multiple calculations of the forward model in the prediction. Other approaches unroll and learn simultaneously the forward and inverse models \cite{liu2021physical}. However, the background medium (the Green's function) needs to be known a priori, and for medical imaging, the properties are not known precisely in advance.

Most of the previous methods reconstruct only one property, such as SoS for US or relative permittivity for radar. Reconstruction of multiple properties is challenging due to the trade-off effects between different parameters and different orders of amplitudes in the wave-field, which make the inversion ill-conditioned \cite{suo2023data}. In addition, some model-based networks were based on a model in the network design which relates only to a specific property such as SoS in \cite{heller2021deep}, where the authors use the coherency measure which needs to be calculated for multiple windows and for each possible discrete SoS. Many techniques also assume only one transmission, which limits the input data and network performance. In the radar domain, previous approaches mainly used a time-harmonic transmission setup, which is not suitable for pulse transmission setups as used in US imaging \cite{wei2018deep, liu2021physical, guo2021physics}. Finally, previous methods were typically tested on simple synthetic tests such as MNIST or various circular shapes \cite{wei2018deep}, and not on realistic clinical settings, or needed specific hardware \cite{jush2020dnn}.

In summary, prior studies utilizing model-based deep learning methods to address the ISP in medical imaging are not applicable to broader transmission setups, such as non-time-harmonic transmissions or setups with elements that do not encompass the objects, such as linear probe setups. Moreover, these previous approaches typically use a large number of elements, often dozens or hundreds, and mainly reconstruct only a single property.

\subsection{Contribution}
\noindent We introduce MB-QRUS, which stands for Model-Based Quantitative Radar and US, a model-based deep learning method for real-time reconstruction of multiple physical properties mappings from either US or radar signals. Our method is based on an unfolding mechanism \cite{shlezinger2023model} of FWI with learned gradients according to a U-Net based block. We use the residuals between the measured CD and the predicted CD, according to physical property estimation and the physical model of wave propagation, to learn the gradients which are often used to update the physical properties estimation. We also introduce a new time-domain and tensor representation of the input measured CD which captures the spatial representation of the CD. We compare our network results to FWI \cite{shultzman2022nonlinear}. To the best of our knowledge, this is the only available method in the literature currently that allows recovering multiple quantitative physical properties mapping, for general transmission setups in US and radar.

Our approach leads to a good reconstruction of two physical properties with lower NRMSE (56\% for US, 67\% for radar), higher SSIM (7.5\% for US, 11\% for radar), and higher PSNR(7.1\% for 433\% for radar), which means the network succeeds to reconstruct the pixels values, besides the shape and positions of the objects. Moreover, our method uses data from only eight elements in contrast to previous methods which need dozens or hundreds of elements. Our network produces real-time results in less than 1 second for complex scenarios including noise in the input CD or nonhomogenous medium background, and realistic data. Finally, our approach allows using diverse transmission setups such as elements that surround the object or a linear probe.

The rest of the paper is organized as follows. Section \ref{problem} formulates the ISP we aim to solve and presents an iterative algorithm based on FWI which is used for comparison. In Section \ref{Method} we present our deep learning approach, based on the wave propagation model, to achieve real-time reconstruction of multiple physical properties from either US or radar signals. Section \ref{results} demonstrates the performance of our method, compared to FWI, for both radar and US signals. Discussion, future directions, and conclusions are presented in Section \ref{ Discussion and Conclusion}.

\section{Problem Formulation and Background}\label{problem}
\subsection{Problem Formulation}
\begin{figure}[!t]
    \centering
    \includegraphics[width=\linewidth]{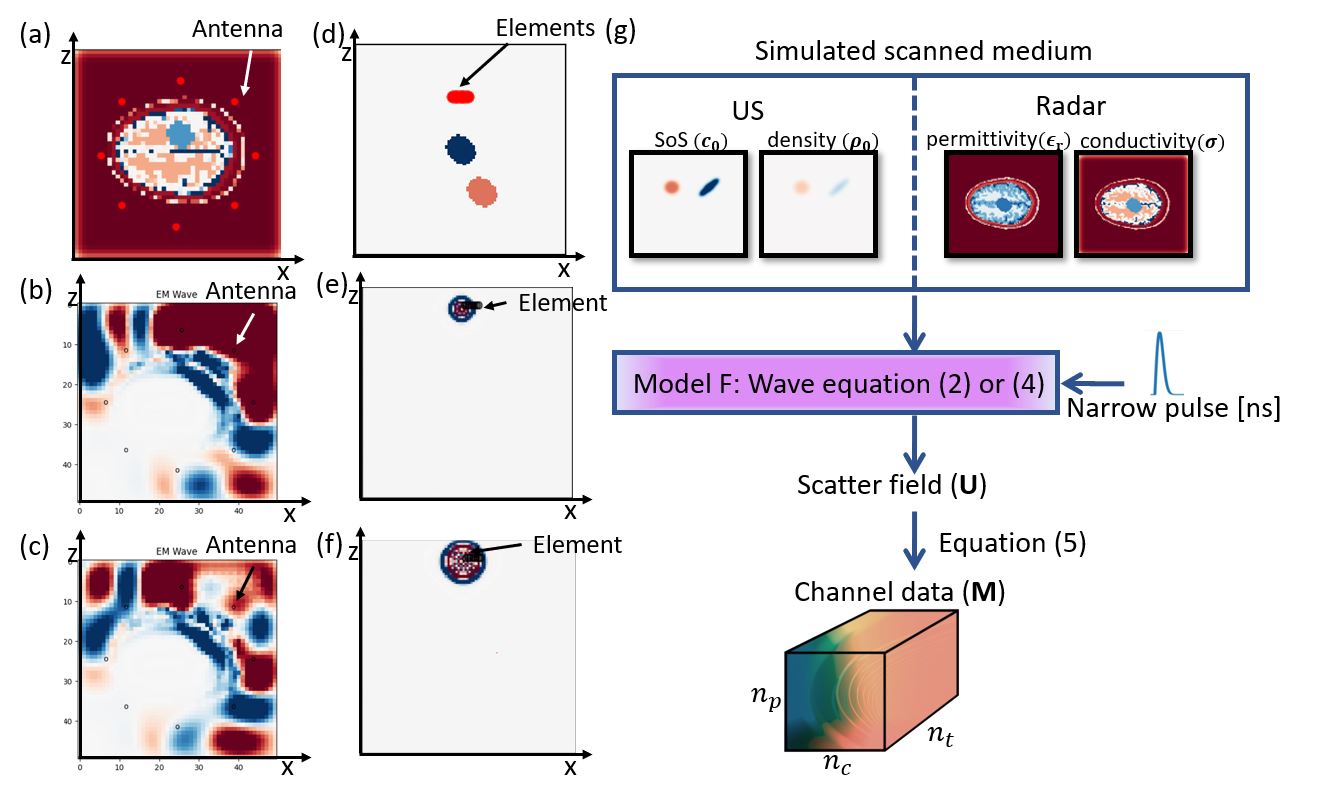}
    \caption{Illustration of the grid setup and data creation. (a)-(c) show an example of the grid setup of a brain with a stroke and 8 antennas surrounding the brain, while (d)-(f) show an example of a US setup with a linear probe and two circles with different physical properties. (b)-(c) demonstrate the wave propagation from one antenna over the grid for two successive time samples, similarly to (e)-(f) for US. (g) displays the creation process of the data, when the CD is created from the simulated physical properties, a known pulse (that defines $\mathbf{S}(t,x,z)$), and utilizing the wave propagation \eqref{eq:us Wave discrete} and, \eqref{eq:radar wave disc}.}
    \label{fig:illustration}
\end{figure}

\noindent We aim to reconstruct multiple physical properties of a scanned medium from radar or US signals. The transmission setup includes $n_c$ elements in a known position (antennas for the radar case and piezoelectric elements for the US case), and $n_p$ known non-interfering pulses, see Fig. \ref{fig:illustration}(a)-(f). We focus on a setup where each element transmits one pulse and the rest of the elements receive the scattered field. The next pulse is transmitted after a time interval of $T$ nano-seconds. This setup is suitable for the transmission process in medical imaging applications that employ radar or US technology \cite{ van1988beamforming, lim2008confocal}.

The scattering field, denoted by $u(t,x,z)$, changes in space and time and is related to the physical properties of the scanned medium by the wave propagation equation. For US, the wave propagation equation is expressed as:
\begin{equation}
\begin{split}
     c_0^2\rho_0\left(\frac{\partial}{\partial x}\left(\frac{1}{\rho_0}\frac{\partial u}{\partial x}\right)+\frac{\partial}{\partial z}\left(\frac{1}{\rho_0}\frac{\partial u}{\partial z}\right)\right) + S = \\ \frac{\partial^2u}{\partial t^2}+2D\frac{\partial u}{\partial t}+D^2u,
\end{split}
\label{eq:US wave}
\end{equation}
where $S(t,x,z)$ represents the source pulse, $c_0(x,z)$ is the SoS of the medium, $\rho_0(x,z)$ is the density of the medium, and $D(x,z)$ is the artificial additional damping term to decrease the needed size of the reconstructed space, called Perfectly Matched Layers (PMLs) \cite{yao2018effective}. For brevity, in \eqref{eq:US wave}, we omitted the brackets in $S$, $c_0$, $\rho_0$ and $D$. 

To obtain a discrete version of the wave propagation equations we use a discrete grid with size $n_x \times n_z$ (Fig. \ref{fig:illustration}(a),(d)) and a discrete form of the time and spatial derivatives \cite{yao2018effective, shultzman2022nonlinear}. The discrete-time derivative is given by a weighted average of the past time samples, and the discrete-spatial derivative is given by a convolution with the Laplacian or gradient kernel. We denote $\textbf{U},\textbf{S}\in \mathbb{R}^{n_x\times n_z\times n_t}$ as the discrete scattering field and source pulse, respectively, where $n_t = \frac{T}{dt}$ and $dt$ is the inverse of the sampling rate. We define, $\mathbf{U}[t]\in \mathbb{R}^{n_x \times n_z}$ as the discrete scattering field for the t'th time step (similarly $\mathbf{S}[t]$). The discrete US wave propagation equation, after organizing the equation such that the scattering $\mathbf{U}[t]$, is dependent on the previous time steps, is given by
\begin{equation}
\begin{split}
      \mathbf{U}[t] = & \Delta^2_t \odot\left[\left(\frac{\mathbf{2}}{\Delta^2_t} -\mathbf{D}^2-\frac{2\mathbf{D}}{\Delta_t}\right)\odot \mathbf{U}[t-1] - \right.\\&\left(\frac{\mathbf{1}}{\Delta^2_t}+\frac{2\mathbf{D}}{\Delta_t}\right)\odot \mathbf{U}[t-2] + \\& \mathbf{C}^2\odot \mathbf{Q} \odot \left(\nabla_D * \left(\frac{1}{\mathbf{Q}}\right)\odot\nabla_D * \mathbf{U}[t-1]\right) + \\& \mathbf{C}^2 \odot \left(\nabla^2_D * \mathbf{U}[t-1]\right) + \mathbf{S}[t]\Big].
\end{split}
\label{eq:us Wave discrete}
\end{equation}

Here $\odot$ is element-wise multiplication, * is the convolution operator, $\nabla_D$ is the discrete gradient filter, $\nabla^2_D$ is the discrete Laplacian filter, 
$\mathbf{1}\in \mathbb{R}^{n_x \times n_z}$ is a matrix of all ones, $\mathbf{2}\in \mathbb{R}^{n_x \times n_z}$ is a matrix of all twos, $\Delta^2_t\in \mathbb{R}^{n_x \times n_z}$ is a matrix with the value dt for each entry, $\mathbf{C}, \mathbf{Q}, \mathbf{D} \in \mathbb{R}^{n_x \times n_z}$ are the discrete SoS, density, and damping, respectively.

For radar signals, we get a similar expression with slight changes due to the electromagnetic wave propagation instead of sound waves:

\begin{equation}
    \nabla^2 u = \frac{\epsilon_r}{c_0^2}\frac{\partial^2 u}{\partial t^2} + \sigma \mu_0 \frac{\partial u}{\partial t} + S,
    \label{eq: radar wave}
\end{equation}
where $\epsilon_r (x,z)$ is the relative permittivity of the medium, $\sigma (x,z)$ is the conductivity of the medium, $c_0$, and $\mu_0$ are the velocity of light and the permeability of the medium (which are constant over the grid), respectively. For brevity, in \eqref{eq: radar wave}, we omitted the brackets in $S, u, c_0, \mu_0, \epsilon_r$ and $\sigma$.

The discrete version of \eqref{eq: radar wave}, similar to the discretization of the wave propagation equation for the US case, is given by:
\begin{equation}
\begin{split}
    &\mathbf{U}[t] = \frac{(\Delta_t \mathbf{C})^2}{\mathbf{\epsilon_r}}\odot \left(\nabla^2_D*\textbf{U}[t-1]\right) + \left(\mathbf{2}-\frac{\mathbf{\sigma}\mu\Delta_t\mathbf{C}^2}{\mathbf{\epsilon_r}}\right) \odot \\ & \textbf{U}[t-1] +  \left(\frac{\mathbf{\sigma}\mu\Delta_t\mathbf{C}^2}{\mathbf{\epsilon_r}}-\mathbf{1}\right) \odot \textbf{U}[t-2]-\left(\frac{(\Delta_t\mathbf{C})^2}{\mathbf{\epsilon_r}}\right)\odot\mathbf{S}[t].
    \label{eq:radar wave disc}
    \end{split}
\end{equation}
Here $ \mathbf{C}\in \mathbb{R}^{n_x \times n_z}$ is a matrix consists of the value $c_o$ in each entry, $\mathbf{\epsilon_r}, \mathbf{\sigma}\in \mathbb{R}^{n_x \times n_z}$ are the discrete relative permittivity and conductivity, respectively.

We define the measured CD for the $p$'th transmission as $\mathbf{M}[p]\in \mathbb{R}^{n_t \times n_c}$ which consists of $n_t$ time samples and $n_c$ receiving channels. It is obtained by a linear mapping, from the scattering field $\mathbf{U}$ using a mapping $\Tilde{R}$ from the spatial signal space to the CD space as follows:
\begin{equation}
\mathbf{M}[p] = \Tilde{R}\mathbf{U}.
\label{eq:cd}
\end{equation}
This mapping can be used for each transmission to obtain the measured CD $\mathbf{M}\in \mathbb{R}^{n_p \times n_t \times n_c}$.

\begin{figure}
    \centering
    \includegraphics[width=\linewidth]{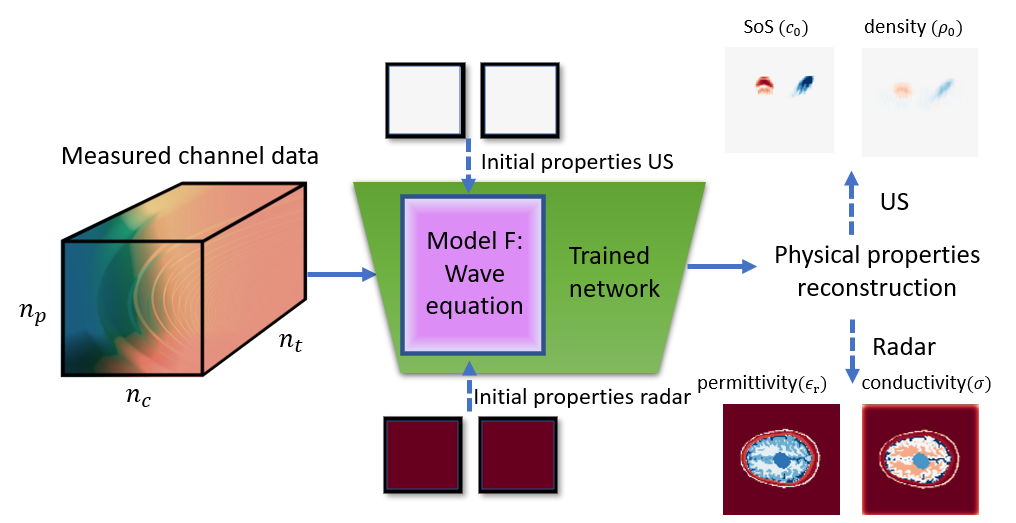}
    \caption{Inference time. The goal of the network is to reconstruct the physical properties mapping from the CD signals.}
    \label{fig:inference}
\end{figure}

Our goal is to reconstruct from the measured CD \textbf{M}, which is related to the scattering field \textbf{U} according to \eqref{eq:cd}, the physical properties mappings $\mathbf{\theta_{j}} \in \mathbb{R}^{n_x \times n_z}$ for $j=1,...,n_m$ where $n_m$ is the number of physical properties. In our case $n_m=2$ and $\mathbf{\theta_j}$ represent the SoS and density for the US case, or relative permittivity and conductivity for the radar case, see Fig. \ref{fig:inference}. 

We aim to design a neural network for this purpose with an unfolding mechanism, that will learn the gradient directions in FWI, based on predicted channel data according to the physical model \eqref{eq:us Wave discrete} or \eqref{eq:radar wave disc}. Using this architecture we not only reconstruct the location and shape of the scanned objects but also achieve precise imaging of their physical properties values.

\subsection{Full Waveform Inversion}
\noindent Full Waveform Inversion (FWI) is an iterative optimization algorithm based on gradient descent to solve the ISP and reconstruct the physical properties mappings from the measured CD, using the knowledge of the wave propagation equations \eqref{eq:us Wave discrete}, \eqref{eq:radar wave disc}. We denote the predicted CD as $\mathbf{\Hat{M}} = F(\{\mathbf{\theta_i}\}_{i=1}^{n_m})$, where $F(\cdot)$ represents the forward wave propagation equations \eqref{eq:us Wave discrete} or  \eqref{eq:radar wave disc}, and $\mathbf{\theta_i}\in \mathbb{R}^{n_x\times n_z}$ is the $i$-th physical property. 
FWI (and similar algorithms such as NWI \cite{shultzman2022nonlinear}) try to minimize the loss between the predicted and measured CD, given by the equation:
\begin{equation}
L = \frac{1}{2} ||\mathbf{\Hat{M}} - \mathbf{M}||_2^2.
\label{eq:FWI}
\end{equation}
The loss function in \eqref{eq:FWI} depends on the predicted physical properties of the scanned medium. In some cases, a regularization term is added to the loss function to improve its performance resulting in:
\begin{equation}
L(\{\theta_i\}_{i=1}^{n_m}) = \frac{1}{2}||M -F(\{\theta_i\}_{i=1}^{n_m})||^2 +\lambda R(\{\theta_i\}_{i=1}^{n_m}).
\label{eq:the_FWI_loss}
\end{equation}
Here, $\{\theta_i\}_{i=1}^{n_m}$ are the predicted physical properties, and $R(\{\theta_i\}_{i=1}^{n_m})$ is the regularization with weight $\lambda$.

At each iteration, the FWI algorithm computes the derivatives of the loss function with respect to the physical properties and uses this information to update the estimation of the physical properties. The derivative calculation can be done using methods explained in \cite{shultzman2022nonlinear}.

FWI based methods are time consuming because of their iterative nature and can take more than 20 minutes and up to hours, depending on the grid size and the CD size. Additionally, FWI algorithms tend to converge to local minima as a result of the high dependency on the choice of the initial guess. The methods also tend to diverge when applied on nonhomogeneous background, meaning when there is large disparity in the physical properties pixels values $\mathbf{\theta}$. This phenomenon is due to the use of the wave propagation PDE model in \eqref{eq:the_FWI_loss} which is sensitive to small changes in the input physical properties \cite{virieux2009overview, fan2022model}. Therefore, these techniques are generally not suitable for medical imaging applications and are currently not implemented in medical imaging systems.

\section{Model-Based Quantitative Radar and Ultrasound}\label{Method}
\begin{figure*}
    \centering
    \includegraphics[width=\textwidth]{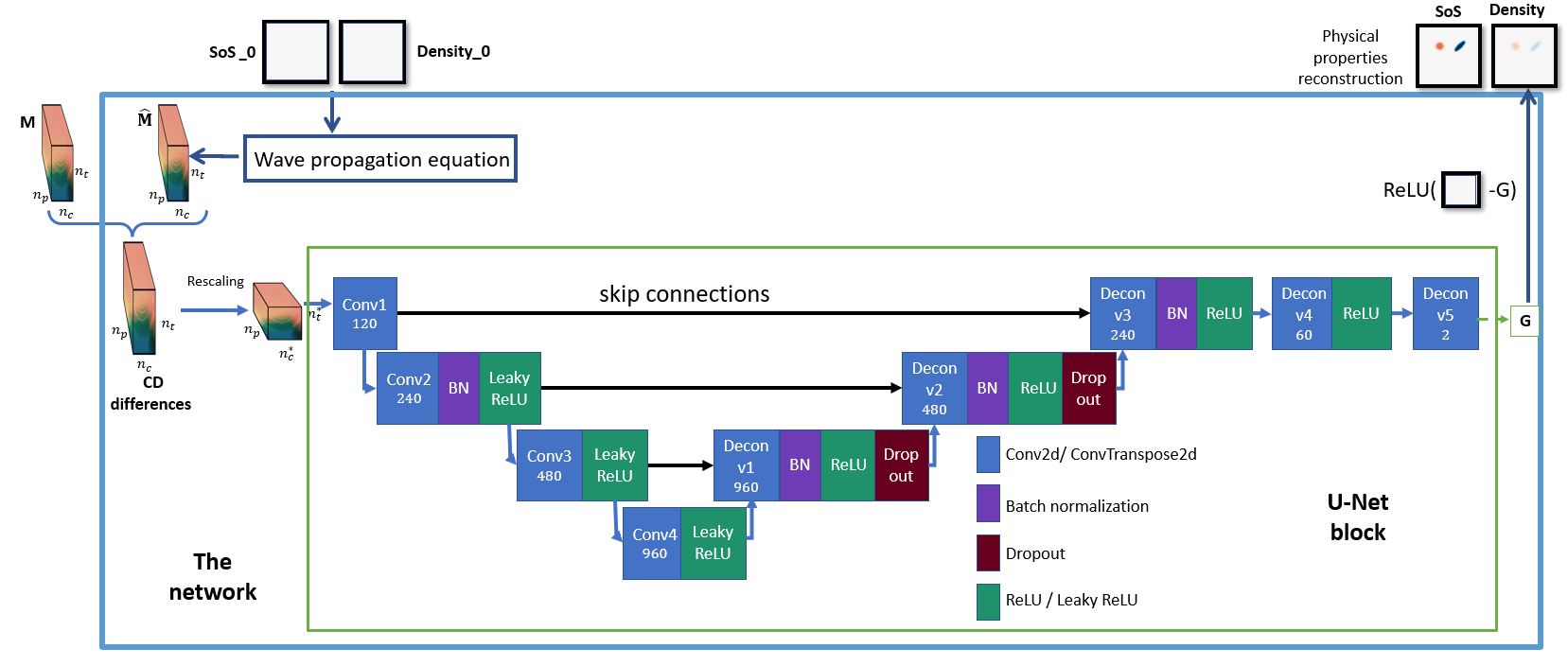}
    \caption{An Example of MB-QRUS architecture for US case. The inputs to the network are initial guesses for the properties and the measured CD. The input to the U-Net block is the CD differences and the output is the gradient tensor $\mathbf{G}$. The output channels number after each convolution is presented.}
    \label{fig:net}
\end{figure*}

\noindent We present MB-QRUS, a model-based deep learning method to reconstruct multiple real-time physical properties mappings from radar or US signals. Our network is designed to learn the gradient $\frac{\partial L}{\partial \theta_i}$ based on the FWI loss \eqref{eq:FWI}. We denote $\textbf{G} = \frac{\partial L}{\partial \theta_i} \in \mathbb{R}^{n_m\times n_x \times n_z}$ and learn it using an U-Net based block, as in Fig. \ref{fig:net}. Our network incorporates in its design the physical model of wave propagation \eqref{eq:us Wave discrete} to calculate a predicted CD $\mathbf{\hat{M}}\in \mathbb{R}^{n_p \times n_t \times n_c}$ and uses the differences between the measured and predicted CD, $\mathbf{M}-\mathbf{\hat{M}}$, as input to the U-Net block to learn $\mathbf{G}$. To calculate the predicted CD, a set of initial mappings $\{\mathbf{\theta}^0_{j}\}_{j=1}^{n_m} \in \mathbb{R}^{n_x \times n_z}$ for $n_m$ physical properties are given as input to our network, as well as the measured CD. The initial guesses were chosen, similarly to FWI initialization, to be the average physical properties of the background medium, which is suitable for medical applications.

By adopting this particular architecture, we have replicated the functionality of known optimization methods such as FWI. However, rather than computing the gradient tensor for the loss during each iteration, which entails calculating the difference $\mathbf{M}-\mathbf{\hat{M}}$, we learned the tensor $\mathbf{G}$ related to this loss, therefore, achieving convergence in fewer steps and data, and to a more accurate solution.

An additional novelty in our architecture is a spatial representation of the CD tensor in the time domain, inspired by FWI. We work with a three-dimensional CD, in contrast to a flattened vector used in previous works, and we utilize the time domain instead of using discrete multi-frequency data as in previous works \cite{fan2022model, wei2018deep, liu2021physical, guo2021physics}. This allows us to refer to the time samples $n_t$ and receiving channels $n_c$ dimensions as spatial dimensions in the convolution layers, whereas the transmission dimension $n_p$ is equivalent to the channels in the convolution layers. The time and receiving channels represent information from different pixels in the grid, therefore referring to them as spatial dimensions enables us to benefit from known methods in convolution networks for images, such as stride convolution for learning the grid properties from the overall CD information.

The U-Net block for learning the tensor $\mathbf{G}$ is composed of stride convolution, batch normalization, dropout layers, and skip connection, each with unique benefits for processing CD for medical imaging. The stride convolutions increase the receptive field of the network, allowing it to have a global view of the input. This enhances the network's ability to learn the grid properties from the overall CD information by analyzing data from non-neighboring receiving channels and time samples. Batch normalization is a powerful technique that addresses the problem of exploding and vanishing gradients when training networks that involve PDEs \cite{nguyen2021parametric}. We used dropout layers with 0.5 probability to avoid overfitting which can occur when the network learns the statistical noise of the training data. Skip connections help the network learn from a global view while preserving fine details and can also mitigate the problem of vanishing gradients, making them beneficial for learning complex relations between input CD and output physical properties mappings. 

The contracting and expansive parts of the U-Net use different activation functions. Leaky ReLU activation is applied in the contracting part to gradually reduce negative values, while ReLU activation is used in the expansive part for the convolution layers. The last convolution layer in the U-Net block has a $1\times1$ kernel to sum over the different transmission channels and output the gradient tensor $\mathbf{G}$ for multiple physical properties. Additionally, We focus on a square grid ($n_x=n_z$), and since $n_t > n_x$ while $n_c < n_x$, we first apply a bilinear interpolation rescaling operator to achieve square spatial dimensions for the input to the U-Net block, which leads to square output spatial dimensions for the gradient tensor $\mathbf{G}$.

The update step for iteration $i+1$ is calculated for each property $\mathbf{\theta_{j}}$ according to:
\begin{equation}
    \mathbf{\theta_{j}^{i+1}} = \mbox{ReLU}(\mathbf{\theta_{j}^{i}}-\mathbf{G_{j}}),
    \label{eq: update network}
\end{equation}
where $\mathbf{G_{j}}\in \mathbb{R}^{n_x \times n_z}$ is the gradient matrix for the j-th physical property. We repeat \eqref{eq: update network} for $L$ layers. We used a loss combined from the Mean-Squared Error (MSE) between the Ground Truth (GT) physical properties and the predicted ones and a Sobel regularization for each predicted property \cite{shultzman2022nonlinear}:
\begin{equation}
    \text{Loss}= \sum_{j=1}^{n_m}\alpha_{j}||\mathbf{\theta}_{j}-\mathbf{\hat{\theta_{j}}}||_2^2 +\beta_j R(\mathbf{\hat{\theta_{j}}}), 
    \label{eq: net_loss}
\end{equation}
where $\mathbf{\theta_{j}}$ and $\mathbf{\hat{\theta_{j}}}$ are the GT and predicted j-th physical property, respectively. Here $R(\mathbf{\hat{\theta_{j}}})$ is a Sobel regularization on the j-th predicted property. A scaling factor for each property to achieve a similar influence effect in the back-propagation process is denoted as $\alpha_j$, and $\beta_j$ is a scaling factor for the influence of the regularization of each property.  Additionally, we normalized the training set of the input-measured CD according to the mean and standard deviation of the flattened vector to achieve better performance in the learning process. The training process is summarized in Algorithm \ref{alg:network} and the inference process is summarized in Algorithm \ref{alg:network_inf}.

\begin{algorithm}[H]
\caption{Training MB-QRUS}
\label{alg:network}
\begin{algorithmic}
\STATE 
\STATE Initialization of UNet weights and biases
\STATE Initialization of $\alpha_j$, $\beta_j$ \hfill for $j=1,...,n_m$
\FOR{epoch=1 to epochs}
\STATE \textbf{Inputs}:
\STATE \hspace{0.5cm} $M \in \mathbb{R}^{n_p \times n_t \times n_c}$ \hfill the measured  CD
\STATE \hspace{0.5cm} $\theta^0_{j} \in \mathbb{R}^{n_x \times n_z}$ for $j=1,...,n_m$ \hfill set of initial properties
\FOR{$i=1$ to $L$}
\STATE $\Hat{M} = F(\{\theta^{i-1}_{j}\}_{j=1}^{n_m})$ \hfill predicted CD from the physical model
\STATE $G = UNet(M-\hat{M})$
\STATE $\theta^{i}_{j} = ReLU(\theta^{i-1}_{j}-G_{j})$ \hfill for $j=1,...,n_m$
\ENDFOR
\STATE Loss = $\sum_{j=1}^{n_m}\alpha_{j}||\mathbf{\theta}^L_{j}-\mathbf{\hat{\theta_{j}}}||_2^2 +\beta_j R(\mathbf{\theta^L_{j}})$ where $\hat{\theta_{j}}$ is the GT property
\STATE Update Unet weights and biases using an optimizer.
\ENDFOR
\STATE \textbf{Output}: trained MB-QRUS model
\end{algorithmic}
\end{algorithm}

\begin{algorithm}[H]
\caption{MB-QRUS inference}
\label{alg:network_inf}
\begin{algorithmic}
\STATE 
\STATE \textbf{Inputs}:
\STATE \hspace{0.5cm} $M \in \mathbb{R}^{n_p \times n_t \times n_c}$ \hfill the measured  CD
\STATE \hspace{0.5cm} $\theta^0_{j} \in \mathbb{R}^{n_x \times n_z}$ for $j=1,...,n_m$ \hfill set of initial properties
\FOR{$i=1$ to $L$}
\STATE $\Hat{M} = F(\{\theta^{i-1}_{j}\}_{j=1}^{n_m})$ \hfill predicted CD from the physical model
\STATE $G = UNet(M-\hat{M})$
\STATE $\theta^{i}_{j} = ReLU(\theta^{i-1}_{j}-G_{j})$ \hfill for $j=1,...,n_m$
\ENDFOR
\STATE \textbf{Output}: $\{\theta^{L}_{j}\}_{j=1}^{n_m}$
\end{algorithmic}
\end{algorithm}

\section{Numerical Results}\label{results}
\noindent In this section we evaluate the performance of our method, using US and radar. The training set consists of normalized CD from 1000 images for each dataset, and the validation set consists of CD from 200 images, normalized according to the mean and variance of the training set. We used the loss function as in \eqref{eq: net_loss} with $n_m=2$. For the US case, $j=1$ and $j=2$ respectively correspond to SoS and density whereas for the radar case $j=1$ and $j=2$ respectively correspond to conductivity and relative permittivity. Additionally $\beta_j = 0$ for the radar scenario. We train all models on a single NVIDIA Quadro RTX8000 GPU with 45GB of memory, and all the experiments are implemented in Pytorch 1.11.0. For both US and radar cases the ADAM optimizer is used, with a learning rate of 0.0001, and a CosineAnnealingLR scheduler with $T_{max} = 20$ and $\eta_{min} = 0$. We use a batch size of 8 for US cases and 16 for radar cases. Each epoch takes approximately 70 seconds for training and we train each network until convergence and without overfitting.

We compare our method to the non-learning optimization based FWI algorithm with loss defined in \eqref{eq:the_FWI_loss} with Sobel regularization operator that enforces soft edges as defined in \cite{shultzman2022nonlinear}. The FWI was initialized with the same values as MB-QRUS of the average scanned medium background. In addition, we used 150 iterations for the algorithm to converge. We did not compare our method to other neural networks because, to the best of our knowledge, there is no such network that can reconstruct multiple properties and is suitable for different transmission setups including a linear probe.

Three different numerical metrics are used to evaluate our method performance, compared to FWI. First we calculate the Normalized Root Mean Squared Error (NRMSE) to evaluate the accuracy of our physical properties reconstruction, following previous works on SoS estimation \cite{shultzman2022nonlinear, rau2021speed}. Second, we examined the Peak Signal-to-Noise Ratio (PSNR) between the reconstructed and GT images to evaluate the quality of the reconstructed properties mapping. Finally, we consider the Structural Similarity Index Measure (SSIM), to evaluate the reconstruction of the shape and size of the scanned objects. The NMRSE is defined as \begin{equation}
    NRMSE\left( {\hat \theta } \right) = \frac{\sqrt{{\left| {\hat \theta } - \theta _{GT} \right|_F^2} /\left( {{n_x}{n_z}} \right)}}{\theta _{\max } - \theta _{\min }},
\end{equation}
where $\theta_{GT}$, $\hat \theta$ are the GT and reconstructed physical properties, respectively. Here $\theta_{\max}$ and $\theta_{\min}$ are the upper and lower bounds on the property values, respectively. The PSNR is defined as
\begin{equation}
    PSNR\left( {\hat \theta } \right) = 20 log_{10}({ \theta_{max} }) - log_{10}||{\hat \theta } - \theta_{GT}||_F^2,
\end{equation}
where $\theta_{max} $ is the maximum physical property value in the GT image. The SSIM is defined as 
\begin{equation}
    SSIM(x,y) = \frac{(2\mu_x\mu_y+c_1)(2\sigma_{xy}+c_2)}{(\mu_x^2+\mu_y^2+c_1)(\sigma_x^2+\sigma_y^2+c_2)},
\end{equation}
where $\mu_x$ is the pixel sample mean of $x$, $\mu_y$ is the pixel sample mean of $y$, $\sigma_x^2$ is the variance of $x$, $\sigma_y^2$ is the variance of $y$, $\sigma_{xy}$ is the covariance of $x$ and $y$, $c_1 = (k_1L)^2$, $c_2=(k_2L)^2$ where $L$ is the dynamic range of the pixels values and $k_1=0.01, k_2=0.03$ by default. 

\subsection{Radar Results}
\noindent For the radar cases, a $30 cm \times 30 cm$ grid is used and discretized into $50 \times 50$ pixels. A PML of 9 pixels is used to prevent reflections from the grid's edges. The permeability is set to $\mu_0 = 1.255\times 10^{-6}$ $\frac{H}{m}$ and the speed of light, $c_0 = 3 \times 10 ^{8}$ $\frac{m}{s}$, 800 times samples are used with dt = 0.005 s and the Courant-Friedrichs-Lew (CFL) \cite{courant1928partiellen} is verified to ensure convergence of the numerical equation to a valid PDE solution.  

We position placed 8 antennas equally on an ellipse, as can be seen in Fig. \ref{fig:illustration}(a). Each antenna emits a Gaussian pulse as the transmission source, with a central frequency $f$ of 1 GHz according to \cite{mobashsher2016design}: 
\begin{equation}
    Src(t) = N\times\sin(2\pi ft)e^{-\frac{2\pi t^2}{0.3^2}},
    \label{eq: source pulse radar}
\end{equation}
with an offset of 10 time samples,

Additionally, the network has only one learned layer ($L=1$), but we repeat the update step \eqref{eq: update network} twice (with the same learned tensor $\mathbf{G}$). We used the MNIST data set \cite{lecun1998gradient} when the digits were placed randomly inside the grid. The digits represented scattered objects that mimic blood with a conductivity of 1.582900 $\frac{S}{m}$ and relative permittivity of 61.065, while the background mimics air with physical properties of conductivity 0.025 $\frac{S}{m}$ and relative permittivity 1.0006. The second dataset for the radar case consists of a simulated real brain slice using MRI scans \cite{qureshi2017levels} generated with a random hemorrhagic stroke, see Fig. \ref{fig:illustration}(g). This complex dataset had a nonhomogeneous background and simulated a real medical application \cite{mobashsher2016design, fhager20193d}. The FWI algorithm and MB-QRUS were initialized with background average values. 

\subsubsection{Brain slices dataset results}
\begin{figure*}
    \centering
    \includegraphics[width=\textwidth]{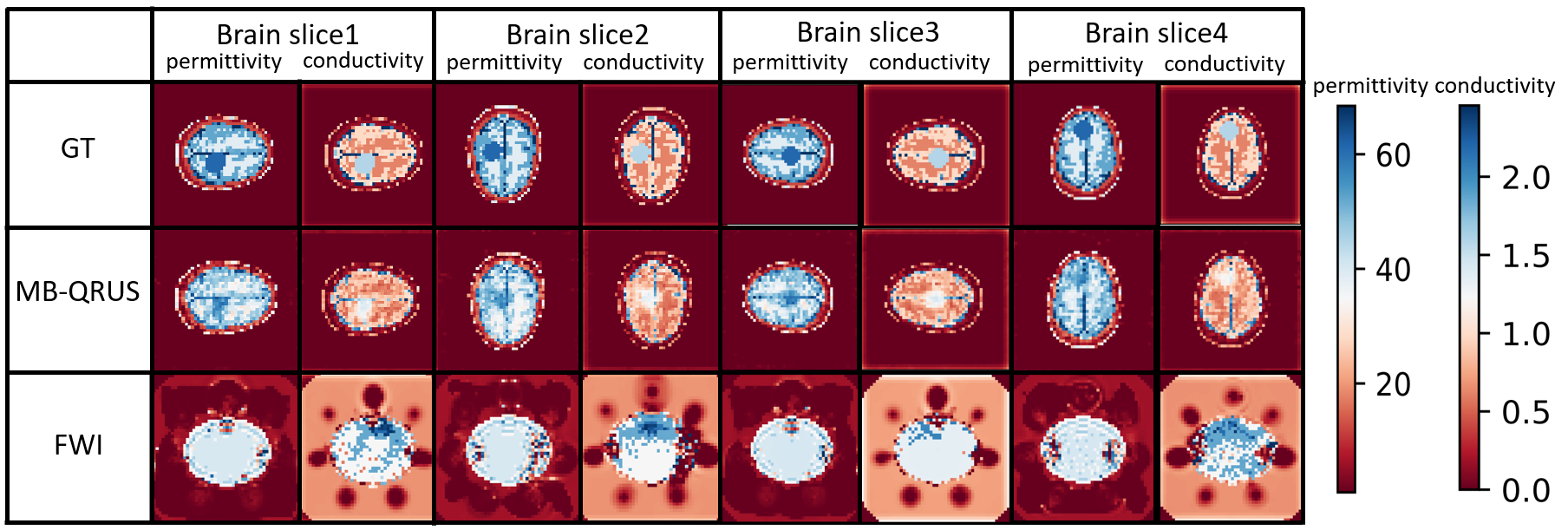}
    \caption{Radar properties reconstruction, by MB-QRUS and FWI compared to the GT for 4 test cases of a realistic brain slice with different orientations and a random stroke.}
    \label{fig:radar_results}
\end{figure*}

Fig. \ref{fig:radar_results} depicts the reconstruction results of MB-QRUS compared to FWI and GT, given 4 test cases of a realistic brain slice with a random stroke and different orientations. Our method successfully reconstructs the stroke position, shape and values, in addition to the brain values and structures, using data from only 8 antennas for different positions and orientations.  The stroke was reconstructed even when it was placed in the middle of the brain and not near the skull, where the skull causes a significant decrease in signal quality. Our approach outperforms FWI for both conductivity and relative permittivity properties for all the cases and metrics. Overall, our method attains lower NRMSE values by 83.53\% for conductivity reconstruction, and by 79.72\% for relative permittivity reconstruction. In addition, our network achieves higher PSNR and SSIM values by 24.61\% and 1150.35\% for the conductivity reconstruction, respectively, and by 3.91\% and 467.33\% for the relative permittivity reconstruction, respectively. Our method achieved better reconstruction of physical properties mappings both in shape, position, and values in less than 0.3 seconds compared to more than 13-31 minutes for the competing FWI method.

\subsubsection{MNIST dataset results}

\begin{figure*}
    \centering
    \includegraphics[width=\textwidth]{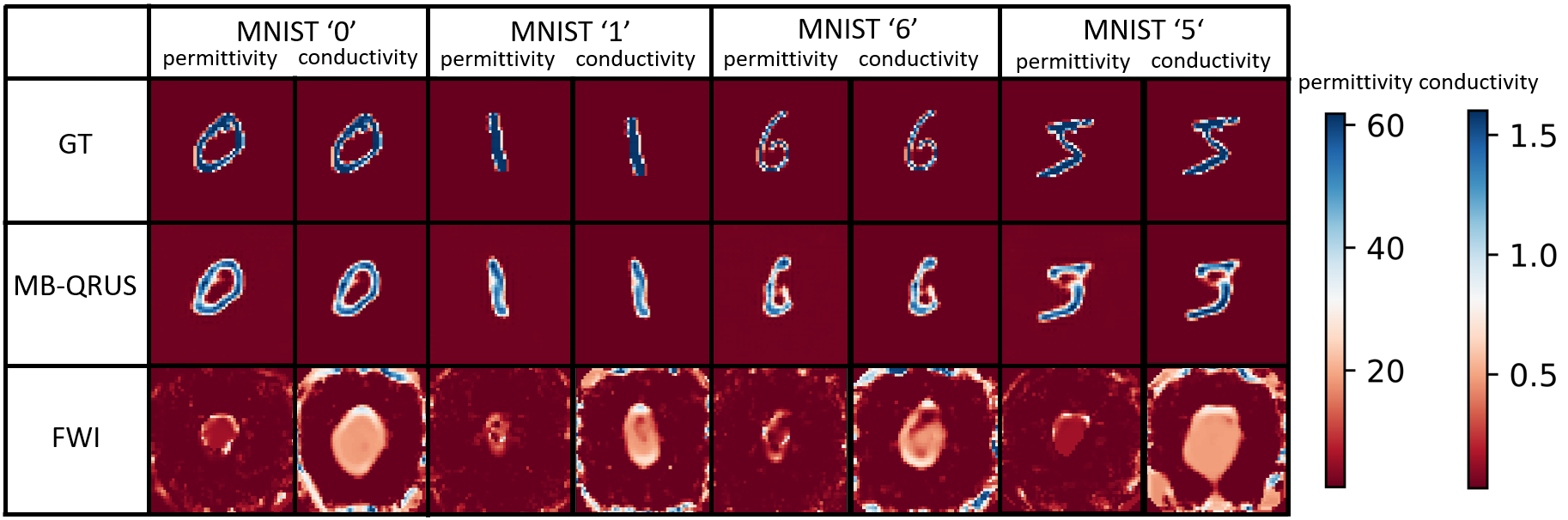}
    \caption{Radar properties reconstruction by MB-QRUS and FWI compared to the GT for 4 cases of a scatter object using MNIST digits shapes (0, 1, 6, and 5), using fixed initialization method. } 
    \label{fig:MNIST_results_radar_fixed}
\end{figure*}

Fig. \ref{fig:MNIST_results_radar_fixed} presents visualization of our method reconstruction compared to FWI and with respect to the GT. We show our network's ability to reconstruct different scatter objects with undefined and complex shapes such as the digits 0, 1, 6 and 5 from data of only 8 antennas. It can be seen in Fig. \ref{fig:MNIST_results_radar_fixed} that the competitive FWI method could not reconstruct any significant results and got only artifacts near the antennas' positions. Our approach takes less than 0.3 seconds to reconstruct the mappings, while FWI takes more than 900 seconds (15 minutes). Overall, our method achieves lower NRMSE values by 62.09\% for conductivity reconstruction, and by 34.19\% for relative permittivity reconstruction. In addition, our network attains higher PSNR and SSIM values by 13.15\% and 1777.58\% for the conductivity reconstruction, respectively, and by 1.17\% and 309.00\% for the relative permittivity reconstruction, respectively.

\subsection{US Results}
\noindent For the US cases, a $5 cm \times 5 cm$ grid is used and discretized into $100 \times 100$ pixels. A PML of size 10 pixels is applied to the grid to prevent reflections from the edges. The CFL \cite{courant1928partiellen} is checked to ensure convergence of the numerical equation to a valid PDE solution, and 240 time samples are used with dt of $1.4077 \times 10^{-7}$ s. 

A simulated transducer array with only 8 elements is used with two different transmission setups: equally spaced elements on an ellipse and a linear probe. The transmission source pulse waveform from each element is a Gaussian pulse centered at the $t_c$ time step, according to the expression:
\begin{equation}
    Src(t) = N \times e^{-f^2((t-t_c)dt)^2}.
    \label{eq: source pulse US}
\end{equation}
The central frequency of the acoustic pulse, $f$, is set to 3 MHz and the shifting of the source time function, $t_c$, is 30 time steps. The pulse amplitude is multiplied by a normalization factor, $N$ which is equal to the inverse of the maximum absolute value of the source pulse, divided by $dt^2$.

We used as a medium synthetic datasets to represent biological organs in terms of size and shape, see Fig. \ref{fig:illustration}(g). The first dataset is of random ovals for a fatty liver scenario. We used physical properties that mimic liver tissue (density of 1060 $\frac{kg}{m^3}$ and a SoS of 1570 $\frac{m}{s}$), while the background properties represent water (density of 1000 $\frac{kg}{m^3}$ and a SoS of 1480 $\frac{m}{s}$). We initialize the FWI algorithm and MB-QRUS with the water values. The network has only one layer ($L$=1). The second dataset is based on MNIST dataset \cite{lecun1998gradient} when the digits are placed randomly inside the grid (without the PML and additional 5 pixels for each direction). The digits shape represents scatter objects that mimic liver tissue with a SoS of 1570 $\frac{m}{s}$ and density of 1060 $\frac{kg}{m^3}$, while the background mimics water with physical properties of SoS of 1480 $\frac{m}{s}$ and density of 1000 $\frac{kg}{m^3}$.
The last dataset is based on segmentation masks derived from authentic CT scans of patients, procured from Kaggle \cite{bilic2023liver}. We choose layers from different patients where there is a liver mask, and then upsample them to be in the grid size. After that we assign a physical properties for water background with SoS of 1480 $\frac{m}{s}$ and density of 1000 $\frac{kg}{m^3}$ and liver object tissue with a SoS of 1570 $\frac{m}{s}$ and density of 1060 $\frac{kg}{m^3}$.

\subsubsection{Random ovals dataset results}

\begin{figure*}
    \centering
    \includegraphics[width=1\textwidth]{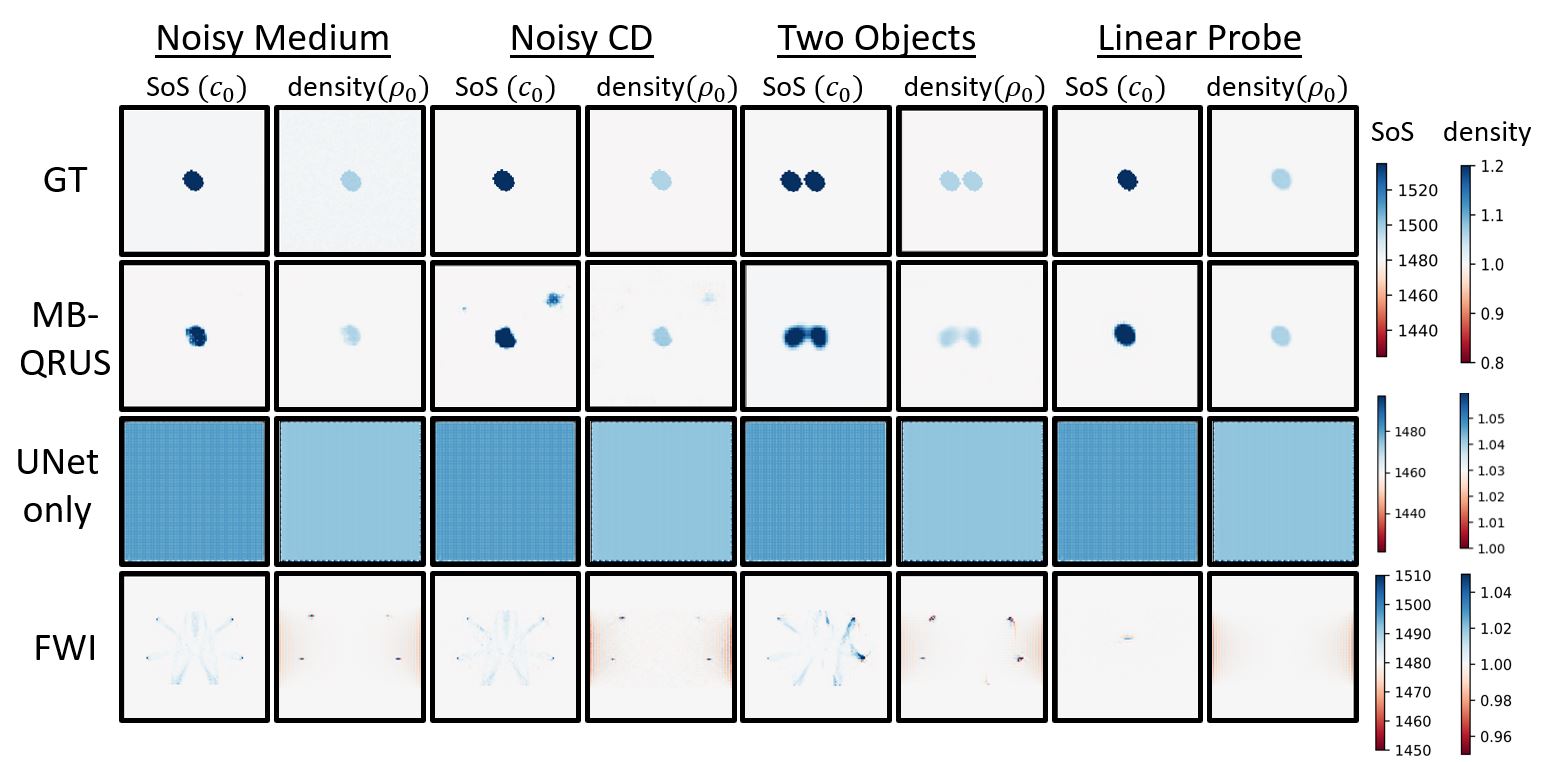}
    \caption{US properties reconstruction, by MB-QRUS compared to GT, reconstruction fron a end-to-end UNet and FWI for 4 test cases: 1. single object - noisy background, 2. noisy CD, 3. two objects, 4. one object for linear probe. For the first 3 cases the network was trained on one random object for each sample with uniform background and elements surrounding the object. For the 4'th case the network was trained with a linear probe setup.}
    \label{fig:US_results}
\end{figure*}

Fig. \ref{fig:US_results} depicts the reconstruction results of our method compared to FWI and GT, given 4 test cases: one object with nonhomogeneous background, noisy CD with additive white noise with a maximum amplitude of 1\% of the maximum value of the signal, 2 objects with uniform background, and one object for a linear probe transmission setup. For the first three cases the network was trained on one random object for each sample and transmission setup of elements that surround the object. The network succeeds to reconstruct the objects from data of only 8 elements even for the nonhomogeneous background or noisy input CD. Moreover, the network was able to generalize and reconstruct two objects even though it was trained on only one object for each sample. Additionally, our network succeeds in reconstructing the object with a more difficult transmission setup of a linear probe when there is no information about the differences between absorption or regular continuous propagation of the signals due to different medium properties. In contrast, FWI was not able to reconstruct any meaningful image due to the small amount of data. Our network outperforms the competitive approach for all the cases, properties, and metrics. Overall, our method attains compared to FWI lower NRMSE values by 56.33\% for SoS reconstruction, and by 55.43\% for density reconstruction. In addition, our network achieves higher PSNR and SSIM values by 1.93\% and 8.15\% for the SoS reconstruction, respectively, and by 13.15\% and 6.10\% for the density reconstruction, respectively.

Our method achieved accurate reconstruction in less than a second compared to more than 32 minutes for FWI.
In addition, we train an end-to-end network consists of only the UNet block part which takes as input the measured CD and output the reconstructed physical properties maps themselves. The training parameters of the end-to-end UNet and MB-QRUS were the same, including the use of data from only 8 elements. The UNet only network wasn't able to learn the complex relation between the CD signals and the physical properties mappings and output similar noise mappings as can be seen in Fig. \ref{fig:US_results}.

\begin{figure}
    \centering
    \includegraphics[width=\linewidth]{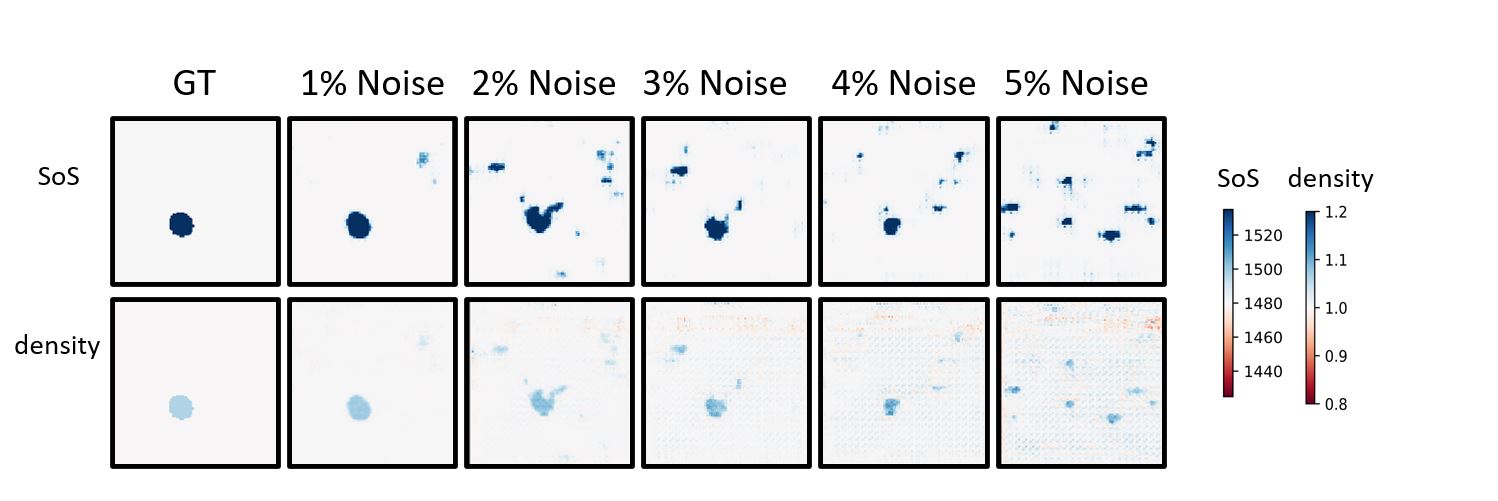}
    \caption{An example of reconstruction SoS and density mapping of random oval with liver properties in a water background with different added noise level to the input CD signals.}
    \label{fig:noise}
\end{figure}

We demonstrate our method ability to reconstruct physical properties mapping, from different level of noise added to the input channel data signal. We added each time a different percentage of the maximum value in the overall channel data tensor to the input signals to the network. This value can be quite large due to recording of the transmission pulse in near by elements. An example of reconstruction results can be found in Fig. \ref{fig:noise} which show a reconstruction results up to 4\% of noise.

\subsubsection{MNIST dataset results}

\begin{figure*}
    \centering
    \includegraphics[width=\textwidth]{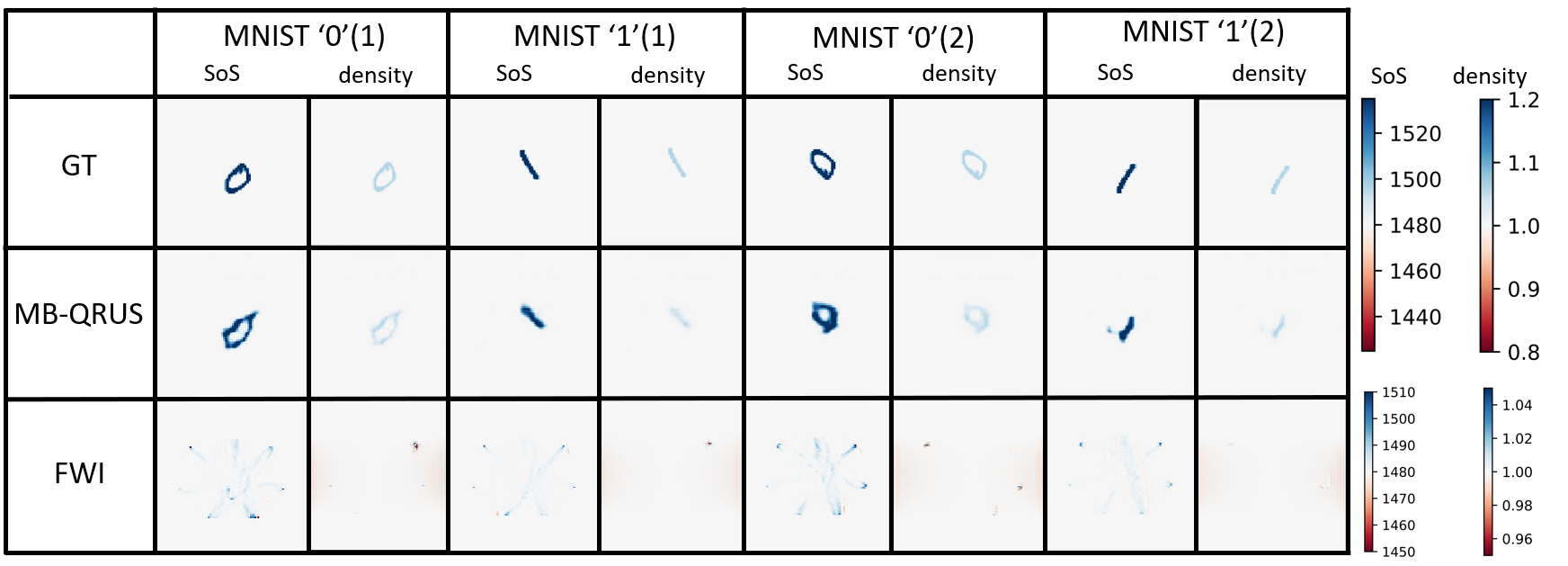}
    \caption{US properties reconstruction by MB-QRUS and FWI compared to the GT, for 4 cases of a scatter object using MNIST digits shapes (2 zeros and 2 ones with different orientations). } 
    \label{fig:MNIST_us_fixed}
\end{figure*}
Fig. \ref{fig:MNIST_us_fixed} demonstrates our network performance compared to FWI with respect to GT. Our method successfully reconstructs undefined shapes that are suitable to describe human organs, using data from only 8 elements. Our approach takes less than 0.2 seconds to reconstruct the mappings, while FWI takes more than 1200 seconds (20 minutes). For the MNIST dataset, we used a transmission setup of elements around the objects. FWI results in artifacts around the positions of the elements as depicted in Fig. \ref{fig:MNIST_us_fixed}. Additionally, it is worth noting the change in scale when comparing the reconstructed values using FWI, which failed to produce meaningful data and instead yielded a grid of pixels with slight perturbations from the initial guesses around the elements' locations. Overall, our method achieves lower NRMSE values by 32.39\% for SoS reconstruction, and by 25.22\% for density reconstruction. In addition, our network attains higher PSNR and SSIM values by 0.56\% and 5.09\% for the SoS reconstruction, respectively, and by 4.14\% and 1.96\% for the density reconstruction, respectively.

\subsubsection{Real liver segmentation maps dataset results}
\begin{figure*}
    \centering
    \includegraphics[width=\textwidth]{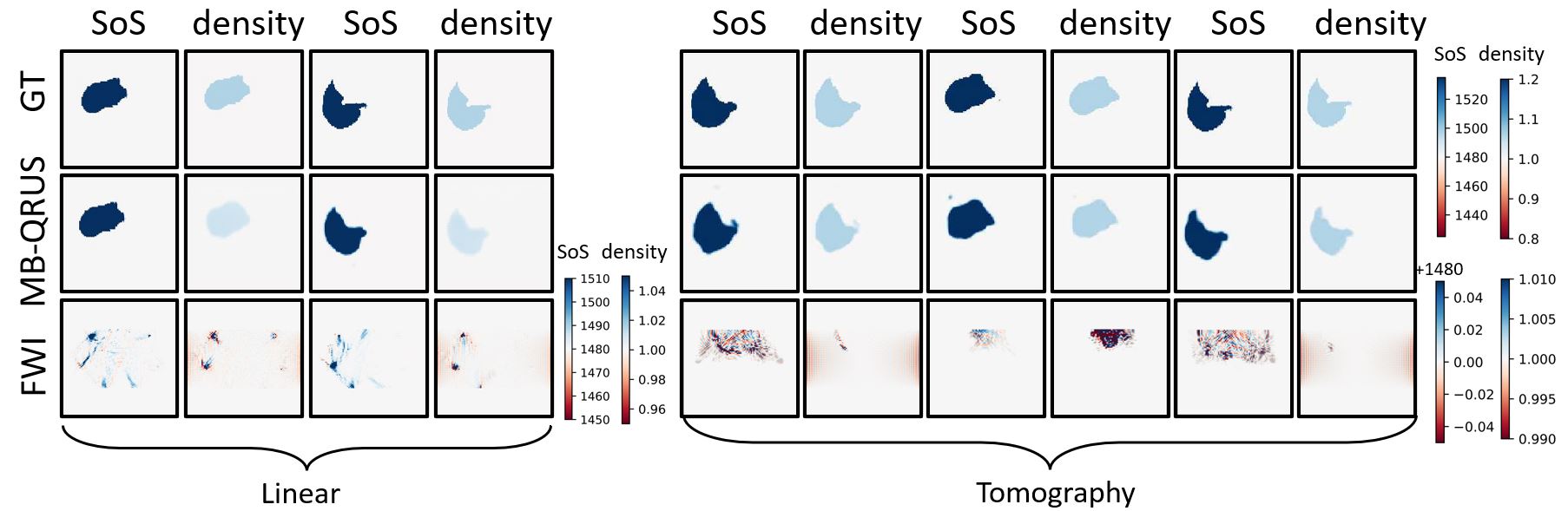}
    \caption{US properties reconstruction by MB-QRUS and FWI compared to the GT, for different realistic livers based on segmentation masks from CT scans. The first two examples from the left are using linear probe transmission setup, and for the rest, the elements are in a circle around the liver.}
    \label{fig:liver_us_mb}
\end{figure*}

We have incorporated a realistic more complex evaluation for the US case, employing liver segmentation masks derived from authentic CT scans of patients, procured from Kaggle \cite{bilic2023liver}. We trained a model using linear probe transmission setup and a model with elements in a circle around the liver, both using only 8 elements. The results show a 65.08\% reduction in the NRMSE, a 1.64\% elevation in the PSNR, and a 12.11\% increase in SSIM for the SoS reconstruction. Correspondingly, the density reconstruction exhibited a reduction of 65.47\% in the NRMSE, a 37.38\% increase in the PSNR, and a 19.68\% increase in SSIM compared to FWI. In addition, we achieved real-time results by less than 0.15 seconds, compared to FWI which took more than 1778 seconds (almost 30 minutes).
A visual comparison of the reconstructed images is presented in Fig. \ref{fig:liver_us_mb}. This dataset provides a realistic scenario for liver scan reconstruction, encompassing both the physical properties' values and the authentic liver shapes. Our network was able to reconstruct the undefined shapes of real patients' livers with high accuracy for both SoS and density mappings, while the FWI did not output any meaningful results and only some artifacts (notice the significant difference in scales).

\subsubsection{Random ovals objects with changing liver values}
\begin{figure*}
    \centering
    \includegraphics[width=\textwidth]{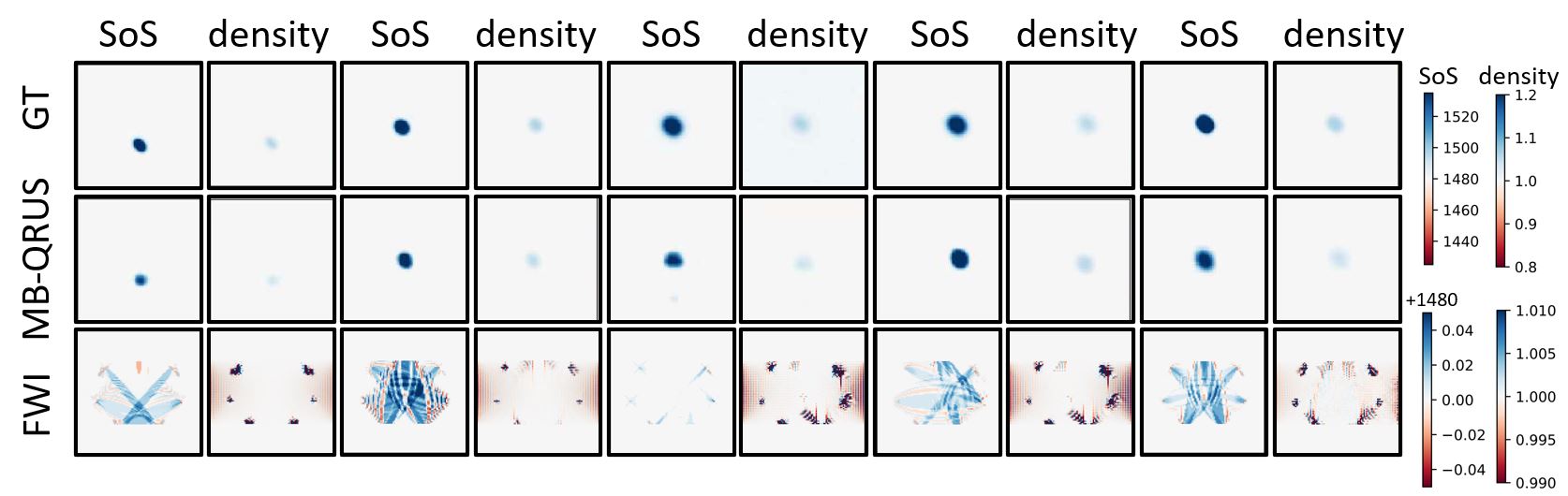}
    \caption{US properties reconstruction by MB-QRUS and FWI compared to the GT, for random ovals with changing liver values distribution.}
    \label{fig:changev}
\end{figure*}

To demonstrate further the reconstruction ability of different changing nonhomogeneous objects values, we create a dataset with random ovals with a distribution of liver values. The object was placed on a water background and was created using a Gaussian filter with different sigma acted on the homogeneous liver properties object, to create a distribution of values around the known average physical values. Fig. \ref{fig:changev} shows examples of the reconstruction results using 8 elements around the object in a circle. From the left, the first two examples were created using a Gaussian filter with sigma of 1.5 for the velocity and 1.8 for the density, and used for inference a model that was trained using the same distributions. The third example from the left, was created  using a Gaussian filter with sigma of 3 for the velocity and density properties, and added Gaussian noise over all the grid of 1\%. A model that was trained using sigma of 1.5 for the velocity and 1.8 for the density properties was used for inference. The forth example from the left, was created using a Gaussian filter with sigma of 3 for the velocity and density, and used for inference a model that was trained using sigma of 1.5 for the velocity and 1.8 for the density properties. The last example from the left was created using a Gaussian filter with sigma of 1.5 for the velocity and 1.8 for the density properties, and used for inference a model that was trained using sigma of 3 for the velocity and density. A transmission setup of 8 elements that surround the object in a circle was used. The results show a 65.10\% reduction in the NRMSE, a 1.57\% elevation in the PSNR, and a 3.59\% increase in SSIM for the SoS reconstruction. Correspondingly, the density reconstruction exhibited a reduction of 66.74\% in the NRMSE, a 18.17\% increase in the PSNR, and a 6.34\% increase in SSIM compared to FWI. In addition, we achieved real-time results by less than 0.15 seconds, compared to FWI which took almost 30 minutes. Our network was able to reconstruct the changing values objects, with different distributions, while the FWI did not output any meaningful results and only some artifacts (notice the significant difference in scales).

\subsection{Real CD from a phantom scan results}
\noindent One of our work's novelties is multiple quantitative imaging results from real measured CD, besides the extensive simulations results. We used a Verasonics research machine for US scans with a P4-2v linear probe to acquire recorded CD from the phantoms scans. It is important to mention that the common US machines that are being used in hospitals today lack the capability to directly access the receiving scattering signals (the CD), only allowing access to the standard B-mode US images after post-processing of the signals. The Verasoncis research platform allows us to acquire and retain the raw CD signals from the US scans for subsequent analysis.  

In the course of our experimentation, we employed a linear probe to scan two different phantoms commonly utilized for human body US research, the 404GS LE and the 403GS LE from Gammex. The phantom background properties are of the average human body commonly used for US scans (SoS of 1540 $\frac{m}{s}$ and density of 1030 $\frac{kg}{m^3}$) and targets of a pin with a diameter of 0.1 mm and physical properties of nylon (SoS of 1070 $\frac{m}{s}$ and density of 1150 $\frac{kg}{m^3}$). One phantom has only one such object, and the second has 3 such nylon objects.  

\begin{figure}
    \centering
    \includegraphics[width=\linewidth]{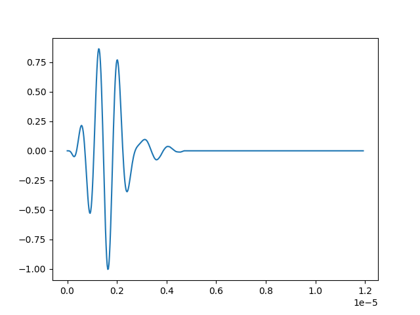}
    \caption{The pulse waveform from Verasonics workspace which is used for the phantom experiment.}
    \label{fig:ver_pulse}
\end{figure}

We scan the phantoms with a transmission setup of Raylines when each element transmits the pulse and the rest record the scattering field. Then we use the data from only the 8 middle elements of the probe as input to the network. We decrease the dt by downsampling the transmission pulse. We used a downsample of factor 2 (denoted as $r$) and we checked the pulse form after the change in sampling rate to ensure the shape wasn't corrupted. Additionally, we adjust all the transmission and network parameters to the ones fitting the Verasoncis scan taken from the scan workspace. These include f = 2.72e6 Hz, T= $\frac{1}{f}$ = 3.6765e-7 s, dt = $\frac{T}{peaks/r} = \frac{T}{16}$ = 2.2978e-08 s, $n_t = 520$, dx= 0.0001 m, the piezoelectric elements positions, and the pulse shape (see Fig. \ref{fig:ver_pulse}. For the dt calculation, we divided $T$ by 16 as we have 16 samples per period. In addition, we check the wave propagation simulation and the CFL condition. Finally, we increase the number of pixels $n_x$ to 200 while $n_z$ is equal to 100 to ensure enough space for the objects' positions.

\begin{figure*}
    \centering
    \includegraphics[width=\textwidth]{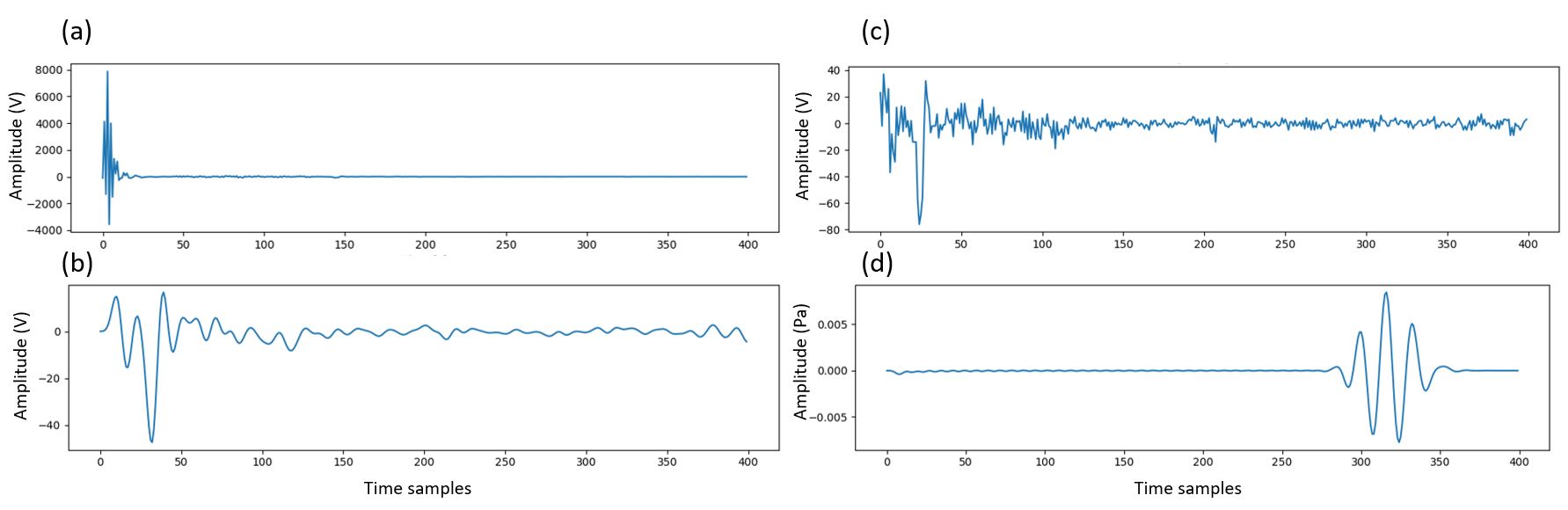}
    \caption{Examples of the real measured CD after the cutting and LPF compared to an example of the simulative one. (a) displays the recorded signal from one element for the first transmission, with the large artifact reflection at the beginning. (b) shows the signal after removing of the first 120 time samples. (c) presents the real measured CD after performing the LPF, and (d) displays an example of the simulative CD for comparison. }
    \label{fig:signals3}
\end{figure*}

There are some main differences when working with the real recorded CD signals compared to the simulative ones. First, there are large reflections at the beginning of the CD time samples from the lenses in the probe and the first entrance to the phantom, see Fig. \ref{fig:signals3}.(a)-(b). These artifacts are not dependent on the object and therefore can be cut from the input CD to the network. It is imperative to bear in mind the necessity of appropriately realigning the reconstruction position in light of this excision. For example, if 120 time samples were removed, with dt = 2.2978e-08 s it means that 2.757e-6 seconds were cut, which are equal to ${C_0}*{T_{cut}} = 0.0042 m$. Given a spatial resolution of dx = 0.0001 m, a consequent displacement of 42 pixels along the Z-axis is mandated for the accurate position reconstruction of the object. It is crucial to note that retrieval of object reconstruction from this specific region is unattainable, as the corresponding data is discarded due to its inherent corruption caused by substantial reflections. Additionally, Verasonic mentions in their manual that any form of image is unachievable within this area

Second, there is much more noise in the real recorded CD signals. To remove the noise, we perform a low-pass filter (LPF). We choose the cutoff frequency to be $3.8 \times 10^6$ Hz and perform the butter LPF with order 6. Fig. \ref{fig:signals3}.(c) presents the filtered signal after the LPF.

Finally, the simulative CD that were used for training and the real measured ones have different ranges of intensity, see Fig. \ref{fig:signals3}.(c)-(d). To overcome this difficulty, we normalized the real recorded CD to be in the same order of ranges as the simulative ones that were used in training.  

\begin{figure*}
    \centering
    \includegraphics[width=\textwidth]{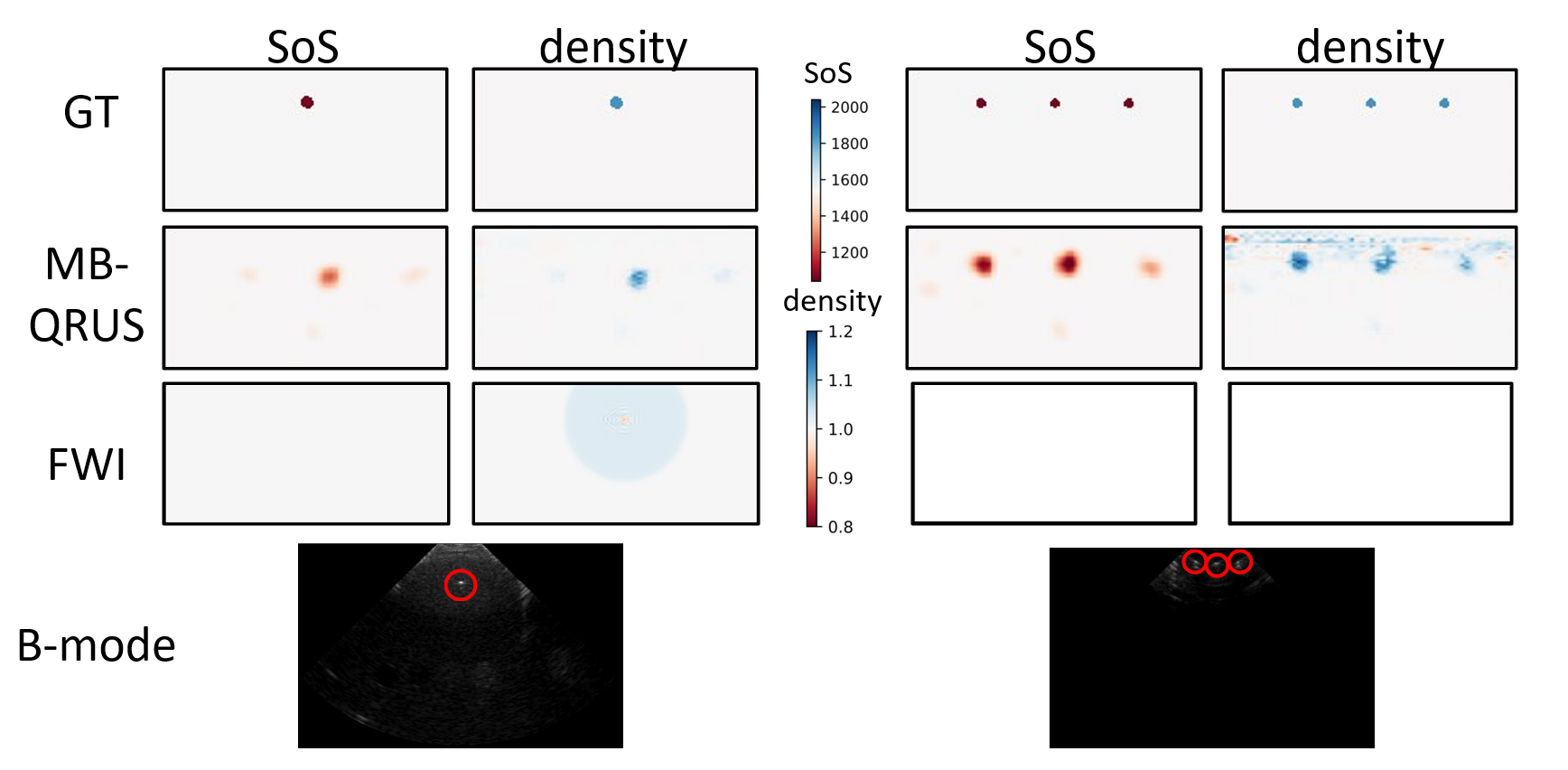}
    \caption{The reconstructed results of SoS and density mappings for two different phantom scans using MB-QRUS and FWI compared to GT. On the bottom row is the B-mode images using Verasonics algorithm for each case. The objects are circled in red.}
    \label{fig:real_data}
\end{figure*}

The reconstruction results using our approach compared to FWI and GT are shown in Fig. \ref{fig:real_data} as a proof-of-concept of our method's performances on real recorded CD. We emphasize a remarkable fact of our findings: our network, trained on data from a single object per example, demonstrated an exceptional capacity to accurately reconstruct the shapes, positions, and properties of three distinct objects within the using real recorded receiving CD signals. This underscores the versatility and robustness of our approach. Additionally, for the second phantom with 3 objects, the competitive FWI approach diverged. At the bottom of Fig. \ref{fig:real_data}, there are the conventional B-mode US images generated by the machine utilizing data from 128 elements. The B-mode images show only the shape and position of the objects and not the different physical properties of the scanned medium, as can be seen using our method, which utilizes data from only 8 elements and not 128. Overall, our method shows promising reconstruction results using real CD and besides the simulative ones. 

\section{Discussion and Conclusion}\label{ Discussion and Conclusion}
\noindent MB-QRUS provides a real-time method for quantitative physical properties imaging from different signals. Our approach integrates the model of wave propagation into the network design to reconstruct mappings from different transmission setups, including the use of a linear probe. Additionally, we utilize a U-Net based block to achieve more accurate values reconstruction for complex scenarios including realistic brain simulations and real measured phantom data. By leveraging the power of wave propagation modeling, spatial time representation, and the U-Net's capabilities, our network allows reconstruction from data of only eight elements. The versatility of our method, including reconstruction from either radar or US signals and advanced transmission setups including a linear probe paves the way for medical quantitative imaging.

Our network mimics the FWI algorithm but learns the gradients from less data. We examine adding to the loss in \eqref{eq: net_loss} also a MSE loss between the measured (input) CD and the predicted one according to the network properties reconstruction and the wave equations (\eqref{eq:us Wave discrete}, \eqref{eq:radar wave disc}), similarly to the FWI loss \eqref{eq:the_FWI_loss}. However, adding this loss to the training process caused a small degradation in the network performance and even a divergence in the learning process. This can be a result of the use of the PDE in the backpropagation due to the calculation of the predicted CD in the loss, which is unstable and can lead to exploding values.

We perform diverse ablation studies as increasing the number of elements used to transmit and record the receiving signals, removing the regularization, or changing the initial guesses type. First, we analyze the impact of initial guess types on the network's performance during both the training and test phases. The fixed initialization in training generally results in sharper reconstructions, particularly noticeable in the case of MNIST digit reconstruction or the stroke itself, irrespective of the initialization type during test time. In addition, the fixed initialization option appears to be more resilient to noise in the CD. However, the random initialization options yield better numerical results in terms of SSIM and PSNR (25.39\% compared to 0.71\% and 29.37\% compared to -6.04\% for density reconstruction, respectively. 316.93\% compared to 39.11\% and 13.15\% compared to -14.88\% for conductivity, respectively. For permittivity, 93.28\% compared to 66.41\% and 1.17\% compared to -0.63\%, respectively. All the percentages are improvements of MB-QRUS results compared to FWI for using the same initialization). 

Additionally, we used the brain slices dataset with radar signal to examine the influence of the number of elements on the results. As the number of elements increased the reconstructed stroke was slightly more defined, and the numerical metrics improved slightly. However, the changes were not significant. Last, regarding Sobel regularization, we noticed slightly improved reconstruction results for the noisy CD case using Sobel regularization. Numeric metrics using US signals, compared to FWI, reveal a 6.85\% SSIM improvement for SoS, offset by minor PSNR and NRMSE reductions (0.13\% and 12.64\% respectively). Density reconstruction benefits moderately from Sobel regularization, showing 1.62\% NRMSE reduction, 0.71\% PSNR increase, and 2.79\% SSIM improvement. We want to mention that the metric results vary for different examples and cases and overall, Sobel regularization contributes slight enhancements but not significantly.

Our method can be extended to a reconstruction of more physical properties, by using a different wave propagation equation that includes those properties. In addition, the method can be extended to the use of CD from different signals, such as seismology and photoacoustics, by adjusting the used wave equations. For future work, we intend to extend our method results using CD from real phantom measurements for complex phantom and scenarios, including using different transmission setups. Additionally, we intend to acquire CD from patients' measurements for diverse applications, for example, patients who suffer from fatty liver disease, and examine our approach.

\bibliographystyle{IEEEtran}
\bibliography{refs}

\begin{thebibliography}{10}
\providecommand{\url}[1]{#1}
\csname url@samestyle\endcsname
\providecommand{\newblock}{\relax}
\providecommand{\bibinfo}[2]{#2}
\providecommand{\BIBentrySTDinterwordspacing}{\spaceskip=0pt\relax}
\providecommand{\BIBentryALTinterwordstretchfactor}{4}
\providecommand{\BIBentryALTinterwordspacing}{\spaceskip=\fontdimen2\font plus
\BIBentryALTinterwordstretchfactor\fontdimen3\font minus \fontdimen4\font\relax}
\providecommand{\BIBforeignlanguage}[2]{{%
\expandafter\ifx\csname l@#1\endcsname\relax
\typeout{** WARNING: IEEEtran.bst: No hyphenation pattern has been}%
\typeout{** loaded for the language `#1'. Using the pattern for}%
\typeout{** the default language instead.}%
\else
\language=\csname l@#1\endcsname
\fi
#2}}
\providecommand{\BIBdecl}{\relax}
\BIBdecl

\bibitem{van1988beamforming}
B.~D. Van~Veen and K.~M. Buckley, ``Beamforming: A versatile approach to spatial filtering,'' \emph{IEEE assp magazine}, vol.~5, no.~2, pp. 4--24, 1988.

\bibitem{lim2008confocal}
H.~B. Lim, N.~T.~T. Nhung, E.-P. Li, and N.~D. Thang, ``Confocal microwave imaging for breast cancer detection: Delay-multiply-and-sum image reconstruction algorithm,'' \emph{IEEE Transactions on Biomedical Engineering}, vol.~55, no.~6, pp. 1697--1704, 2008.

\bibitem{ruby2019breast}
L.~Ruby, S.~J. Sanabria, K.~Martini, K.~J. Dedes, D.~Vorburger, E.~Oezkan, T.~Frauenfelder, O.~Goksel, and M.~B. Rominger, ``Breast cancer assessment with pulse-echo speed of sound ultrasound from intrinsic tissue reflections: Proof-of-concept,'' \emph{Investigative radiology}, vol.~54, no.~7, pp. 419--427, 2019.

\bibitem{lee2017imaging}
D.~H. Lee, ``Imaging evaluation of non-alcoholic fatty liver disease: focused on quantification,'' \emph{Clinical and molecular hepatology}, vol.~23, no.~4, p. 290, 2017.

\bibitem{hopfer2017electromagnetic}
M.~Hopfer, R.~Planas, A.~Hamidipour, T.~Henriksson, and S.~Semenov, ``Electromagnetic tomography for detection, differentiation, and monitoring of brain stroke: A virtual data and human head phantom study.'' \emph{IEEE Antennas and Propagation Magazine}, vol.~59, no.~5, pp. 86--97, 2017.

\bibitem{ireland2011microwave}
D.~Ireland and M.~Bialkowski, ``Microwave head imaging for stroke detection,'' \emph{Progress In Electromagnetics Research M}, vol.~21, pp. 163--175, 2011.

\bibitem{wei2018deep}
Z.~Wei and X.~Chen, ``Deep-learning schemes for full-wave nonlinear inverse scattering problems,'' \emph{IEEE Transactions on Geoscience and Remote Sensing}, vol.~57, no.~4, pp. 1849--1860, 2018.

\bibitem{shultzman2022nonlinear}
A.~Shultzman and Y.~C. Eldar, ``Nonlinear waveform inversion for quantitative ultrasound,'' \emph{IEEE transactions on computational imaging}, vol.~8, pp. 893--904, 2022.

\bibitem{guasch2020full}
L.~Guasch, O.~Calder{\'o}n~Agudo, M.-X. Tang, P.~Nachev, and M.~Warner, ``Full-waveform inversion imaging of the human brain,'' \emph{NPJ digital medicine}, vol.~3, no.~1, p.~28, 2020.

\bibitem{virieux2009overview}
J.~Virieux and S.~Operto, ``An overview of full-waveform inversion in exploration geophysics,'' \emph{Geophysics}, vol.~74, no.~6, pp. WCC1--WCC26, 2009.

\bibitem{guo2023physics}
R.~Guo, T.~Huang, M.~Li, H.~Zhang, and Y.~C. Eldar, ``Physics-embedded machine learning for electromagnetic data imaging: Examining three types of data-driven imaging methods,'' \emph{IEEE Signal Processing Magazine}, vol.~40, no.~2, pp. 18--31, 2023.

\bibitem{chen2020review}
X.~Chen, Z.~Wei, M.~Li, and P.~Rocca, ``A review of deep learning approaches for inverse scattering problems (invited review),'' \emph{Progress In Electromagnetics Research}, vol. 167, pp. 67--81, 2020.

\bibitem{shlezinger2022model}
N.~Shlezinger, Y.~C. Eldar, and S.~P. Boyd, ``Model-based deep learning: On the intersection of deep learning and optimization,'' \emph{IEEE Access}, vol.~10, pp. 115\,384--115\,398, 2022.

\bibitem{shlezinger2023model}
N.~Shlezinger, J.~Whang, Y.~C. Eldar, and A.~G. Dimakis, ``Model-based deep learning,'' \emph{Proceedings of the IEEE}, 2023.

\bibitem{li2018deepnis}
L.~Li, L.~G. Wang, F.~L. Teixeira, C.~Liu, A.~Nehorai, and T.~J. Cui, ``Deepnis: Deep neural network for nonlinear electromagnetic inverse scattering,'' \emph{IEEE Transactions on Antennas and Propagation}, vol.~67, no.~3, pp. 1819--1825, 2018.

\bibitem{jin2020physics}
Y.~Jin, Q.~Shen, X.~Wu, J.~Chen, and Y.~Huang, ``A physics-driven deep-learning network for solving nonlinear inverse problems,'' \emph{Petrophysics}, vol.~61, no.~01, pp. 86--98, 2020.

\bibitem{liu2021ultrasound}
X.~Liu and M.~Almekkawy, ``Ultrasound computed tomography using physical-informed neural network,'' in \emph{2021 IEEE International Ultrasonics Symposium (IUS)}.\hskip 1em plus 0.5em minus 0.4em\relax IEEE, 2021, pp. 1--4.

\bibitem{zhang2020adjoint}
W.~Zhang, J.~Gao, Z.~Gao, and H.~Chen, ``Adjoint-driven deep-learning seismic full-waveform inversion,'' \emph{IEEE Transactions on Geoscience and Remote Sensing}, vol.~59, no.~10, pp. 8913--8932, 2020.

\bibitem{fan2022model}
Y.~Fan, H.~Wang, H.~Gemmeke, T.~Hopp, and J.~Hesser, ``Model-data-driven image reconstruction with neural networks for ultrasound computed tomography breast imaging,'' \emph{Neurocomputing}, vol. 467, pp. 10--21, 2022.

\bibitem{prasad2022deepuct}
S.~Prasad and M.~Almekkawy, ``Deepuct: Complex cascaded deep learning network for improved ultrasound tomography,'' \emph{Physics in Medicine \& Biology}, vol.~67, no.~6, p. 065008, 2022.

\bibitem{rao2020multi}
J.~Rao, J.~Yang, M.~Ratassepp, and Z.~Fan, ``Multi-parameter reconstruction of velocity and density using ultrasonic tomography based on full waveform inversion,'' \emph{Ultrasonics}, vol. 101, p. 106004, 2020.

\bibitem{suo2023data}
M.~Suo, D.~Zhang, H.~Yang, and Y.~Yang, ``Data-driven full waveform inversion for ultrasonic bone quantitative imaging,'' \emph{Neural Computing and Applications}, pp. 1--17, 2023.

\bibitem{jin2021deep}
Y.~Jin, Y.~Zi, W.~Hu, X.~Wu, and J.~Chen, ``A deep learning enhanced full waveform inversion scheme,'' in \emph{2021 International Applied Computational Electromagnetics Society Symposium (ACES)}.\hskip 1em plus 0.5em minus 0.4em\relax IEEE, 2021, pp. 1--4.

\bibitem{liu2021physical}
J.~Liu, H.~Zhou, T.~Ouyang, Q.~Liu, and Y.~Wang, ``Physical model-inspired deep unrolling network for solving nonlinear inverse scattering problems,'' \emph{IEEE transactions on antennas and propagation}, vol.~70, no.~2, pp. 1236--1249, 2021.

\bibitem{guo2021physics}
R.~Guo, Z.~Lin, T.~Shan, X.~Song, M.~Li, F.~Yang, S.~Xu, and A.~Abubakar, ``Physics embedded deep neural network for solving full-wave inverse scattering problems,'' \emph{IEEE transactions on antennas and propagation}, 2021.

\bibitem{ren2020physics}
Y.~Ren, X.~Xu, S.~Yang, L.~Nie, and Y.~Chen, ``A physics-based neural-network way to perform seismic full waveform inversion,'' \emph{IEEE Access}, vol.~8, pp. 112\,266--112\,277, 2020.

\bibitem{hu2021theory}
Y.~Hu, Y.~Jin, X.~Wu, and J.~Chen, ``A theory-guided deep neural network for time domain electromagnetic simulation and inversion using a differentiable programming platform,'' \emph{IEEE Transactions on Antennas and Propagation}, vol.~70, no.~1, pp. 767--772, 2021.

\bibitem{heller2021deep}
M.~Heller and G.~Schmitz, ``Deep learning-based speed-of-sound reconstruction for single-sided pulse-echo ultrasound using a coherency measure as input feature,'' in \emph{2021 IEEE International Ultrasonics Symposium (IUS)}.\hskip 1em plus 0.5em minus 0.4em\relax IEEE, 2021, pp. 1--4.

\bibitem{jush2020dnn}
F.~K. Jush, M.~Biele, P.~M. Dueppenbecker, O.~Schmidt, and A.~Maier, ``Dnn-based speed-of-sound reconstruction for automated breast ultrasound,'' in \emph{2020 IEEE International Ultrasonics Symposium (IUS)}.\hskip 1em plus 0.5em minus 0.4em\relax IEEE, 2020, pp. 1--7.

\bibitem{yao2018effective}
G.~Yao, N.~V. Da~Silva, and D.~Wu, ``An effective absorbing layer for the boundary condition in acoustic seismic wave simulation,'' \emph{Journal of Geophysics and Engineering}, vol.~15, no.~2, pp. 495--511, 2018.

\bibitem{nguyen2021parametric}
V.~M. Nguyen-Thanh, C.~Anitescu, N.~Alajlan, T.~Rabczuk, and X.~Zhuang, ``Parametric deep energy approach for elasticity accounting for strain gradient effects,'' \emph{Computer Methods in Applied Mechanics and Engineering}, vol. 386, p. 114096, 2021.

\bibitem{rau2021speed}
R.~Rau, D.~Schweizer, V.~Vishnevskiy, and O.~Goksel, ``Speed-of-sound imaging using diverging waves,'' \emph{International journal of computer assisted radiology and surgery}, vol.~16, no.~7, pp. 1201--1211, 2021.

\bibitem{courant1928partiellen}
R.~Courant, K.~Friedrichs, and H.~Lewy, ``{\"U}ber die partiellen differenzengleichungen der mathematischen physik,'' \emph{Mathematische annalen}, vol. 100, no.~1, pp. 32--74, 1928.

\bibitem{mobashsher2016design}
A.~Mobashsher, K.~Bialkowski, A.~Abbosh, and S.~Crozier, ``Design and experimental evaluation of a non-invasive microwave head imaging system for intracranial haemorrhage detection,'' \emph{Plos one}, vol.~11, no.~4, p. e0152351, 2016.

\bibitem{lecun1998gradient}
Y.~LeCun, L.~Bottou, Y.~Bengio, and P.~Haffner, ``Gradient-based learning applied to document recognition,'' \emph{Proceedings of the IEEE}, vol.~86, no.~11, pp. 2278--2324, 1998.

\bibitem{qureshi2017levels}
A.~M. Qureshi and Z.~Mustansar, ``Levels of detail analysis of microwave scattering from human head models for brain stroke detection,'' \emph{PeerJ}, vol.~5, p. e4061, 2017.

\bibitem{fhager20193d}
A.~Fhager, S.~Candefjord, M.~Elam, and M.~Persson, ``3d simulations of intracerebral hemorrhage detection using broadband microwave technology,'' \emph{Sensors}, vol.~19, no.~16, p. 3482, 2019.

\bibitem{bilic2023liver}
P.~Bilic, P.~Christ, H.~B. Li, E.~Vorontsov, A.~Ben-Cohen, G.~Kaissis, A.~Szeskin, C.~Jacobs, G.~E.~H. Mamani, G.~Chartrand \emph{et~al.}, ``The liver tumor segmentation benchmark (lits),'' \emph{Medical Image Analysis}, vol.~84, p. 102680, 2023.

\end{thebibliography}
\end{document}